\newif\ifieeeclass
\IfFileExists{IEEEtran.cls}{\ieeeclasstrue\documentclass[conference]{IEEEtran}}{\ieeeclassfalse\documentclass[11pt,twocolumn]{article}}
\ifieeeclass\else
\usepackage[a4paper,margin=0.75in]{geometry}
\newcommand{\IEEEauthorblockN}[1]{#1}
\newenvironment{IEEEkeywords}{\noindent\textbf{Index Terms}---}{\par}
\fi
\usepackage{amsmath,amssymb}
\usepackage{graphicx}
\usepackage{booktabs}
\usepackage{hyperref}
\usepackage{siunitx}
\usepackage{float}
\floatstyle{ruled}
\newfloat{algorithm}{t}{loa}
\floatname{algorithm}{Algorithm}
\newif\ifanonymous
\anonymousfalse
\title{Reward--Modulated Local Learning in Spiking Encoders:\\Controlled Benchmarks with STDP and Hybrid Rate Readouts}
\ifanonymous
\author{\IEEEauthorblockN{Anonymous Submission}}
\else
\author{\IEEEauthorblockN{Debjyoti Chakraborty}\\Researcher\\spoutop@gmail.com}
\fi
\date{}

\begin{document}
\maketitle

\begin{abstract}
This paper presents a controlled empirical study of biologically motivated local learning for handwritten digit recognition. We evaluate an STDP-inspired competitive proxy and a practical hybrid benchmark built on the same spiking population encoder. The proxy is motivated by leaky integrate and fire E/I circuit models with three-factor delayed reward modulation. The hybrid update is local in pre$\times$post rates but uses supervised labels and no timing-based credit assignment. On \texttt{sklearn} digits, fixed-seed evaluation shows classical pixel baselines from 98.06 to 98.22\% accuracy, while local spike-based models reach 86.39$\pm$4.75\% (hybrid default) and 87.17$\pm$3.74\% (STDP-style competitive proxy). Ablations identify normalization and reward-shaping settings as the strongest observed levers, with a best hybrid ablation of 95.52$\pm$1.11\%. A network-free synthetic temporal benchmark supports the same timing-versus-rate interpretation under matched local-update training. A descriptive 2$\times$2 analysis further shows reward-shaping effects can reverse sign across stabilization regimes, so reward-shaping conclusions should be reported jointly with normalization settings.
\end{abstract}

\section{Introduction}
Deep networks usually rely on gradient backpropagation and global error signals. In contrast, cortical learning is often modeled with local plasticity plus neuromodulatory gating. STDP~\cite{bi1998hebbian,gerstner2002spiking} captures timing-dependent synaptic change, while dopamine-like reward can serve as a third factor linking delayed outcomes to local eligibility traces~\cite{izhikevich2007solving}. This paper evaluates that paradigm in a controlled setting by separating timing-based reward-modulated plasticity from a practical local rate-readout benchmark built on the same spiking encoder.

\section{Related work and positioning}
\subsection{Reward-Modulated Local Plasticity}
Three-factor learning rules that combine local pre/post activity with global reward or neuromodulatory signals are a central line of biologically plausible learning~\cite{florian2007reinforcement,fremaux2010functional,izhikevich2007solving}. These works motivate the eligibility-trace formulation used here.

\subsection{Supervised SNN Training}
Modern SNN accuracy improvements are often driven by gradient-based training, including direct backpropagation variants and surrogate gradients~\cite{lee2016training,wu2018spatio,neftci2019surrogate,eshraghian2023training}. Such methods typically outperform strictly local STDP pipelines on benchmark accuracy, at reduced biological locality. We intentionally avoid surrogate-gradient training in this study to preserve strict local-update assumptions and isolate reward-modulated local design effects.

\subsection{Unsupervised STDP and Hardware Perspectives}
Unsupervised STDP systems can learn competitive representations in constrained settings~\cite{diehl2015unsupervised}, while neuromorphic hardware efforts such as TrueNorth and Loihi demonstrate strong potential for low-power event-driven computation and on-chip learning~\cite{merolla2014million,davies2018loihi}. Broader reviews discuss the trade-offs among biological plausibility, trainability, and deployment efficiency~\cite{pfeiffer2018deep,bellec2020solution}.

\subsection{Positioning of This Work}
We position this work as an empirical \emph{local-learning study}, not an accuracy state-of-the-art claim. The main contribution is a controlled and reproducible benchmark protocol that isolates local design choices (normalization schedule, reward shaping, and encoding settings) under fixed seeds and no retuning. We treat the three-factor STDP equations as standard theoretical motivation and focus novelty on interaction-aware ablation evidence, split-robust effect direction, and artifact-backed reproducibility.
Concretely, our novelty thesis has three parts: (i) a controlled benchmark methodology for local-learning variants on a fixed spiking encoder with fixed seeds, split robustness, and scripted regeneration; (ii) an interaction result where stabilization schedule dominates and modulates reward-shaping direction; and (iii) a diagnostic principle showing timing-coded tasks break count readouts unless timing-aware readouts are used.

\section{Problem framing}
This work studies static digit classification as a controlled testbed for local-learning design choices in spiking encoders. The primary evaluation target is not leaderboard accuracy but effect direction and stability across fixed seeds and split protocols. We compare two evaluated branches under shared encoding: a practical local rate readout and an STDP-inspired competitive proxy. Key outcomes are (i) which factors dominate variance, (ii) whether those effects persist across split seeds and external checks, and (iii) whether timing-sensitive tasks expose count-readout limitations.

\section{Contributions}
\begin{itemize}
  \item A reproducible fixed-seed benchmark protocol for local-learning comparisons, including controlled splits, deterministic seeded stochastic sampling, and artifact-backed table/figure generation.
  \item A proxy-scoped evaluation of STDP-inspired competitive updates and a practical local rate-readout branch under a shared encoder, with explicit separation between theoretical motivation and evaluated implementation.
  \item Interaction-aware ablation evidence that identifies dominant normalization/reward-shaping effects, validates direction across split seeds, and contextualizes timing limitations with synthetic temporal and external OpenML checks.
\end{itemize}

\section{Background}
\paragraph{Spike generation.} Static inputs are encoded as Poisson spike trains. For each feature value $x\in[0,1]$ we create a small population of $K$ neurons with Gaussian tuning curves centered at $\{\mu_k\}_{k=1..K}$. The instantaneous firing probability of neuron $k$ at time bin $\Delta t$ is
\begin{equation}
\lambda_k(x) = \lambda_{\max} \exp\Big(-\tfrac{(x-\mu_k)^2}{2\sigma^2}\Big), \qquad p_k = \lambda_k(x)\,\Delta t.
\end{equation}

\paragraph{LIF dynamics.} Neuron membrane $v$ accumulates synaptic current $I$ with leak:
\begin{align}
v(t+\Delta t)&=\alpha v(t) + (1-\alpha)I(t) + I_0,\\
\alpha&=\exp(-\Delta t/\tau_\mathrm{m}).
\end{align}
When $v\geq v_\theta$ the neuron emits a spike and is reset for a refractory period.

\paragraph{Three--factor STDP.} Let $x_i(t)$ and $y_j(t)$ be pre/post spikes. We maintain pre/post traces $\hat x_i,\hat y_j$ and a synaptic eligibility trace $e_{ij}$:
\begin{align}
\hat x_i(t+\Delta t) &= \beta\,\hat x_i(t) + x_i(t),\\
\hat y_j(t+\Delta t) &= \gamma\,\hat y_j(t) + y_j(t),\\
e_{ij}(t+\Delta t) &= \delta\,e_{ij}(t) + A_+\,\hat x_i(t)\,y_j(t) \nonumber\\
&\quad - A_-\,x_i(t)\,\hat y_j(t),
\end{align}
where $\beta=\exp(-\Delta t/\tau_\mathrm{pre})$, $\gamma=\exp(-\Delta t/\tau_\mathrm{post})$, and $\delta=\exp(-\Delta t/\tau_\mathrm{elig})$. A delayed dopamine signal $R(t)$ converts $e_{ij}$ into persistent change:
\begin{equation}
\Delta w_{ij} = \eta\, R(t)\, e_{ij}(t), \qquad w_{\min} \le w_{ij} \le w_{\max}.
\end{equation}
These equations define the motivating local-learning formalism. The results in this paper evaluate the competitive proxy in Algorithm~\ref{alg:proxy_update}, which preserves local reward-shaped competition and bounded updates while abstracting away full conductance-level recurrent dynamics.
\paragraph{\textit{Evaluated artifact statement.}}
\textit{Evaluated artifact in this paper: Algorithm~\ref{alg:proxy_update} competitive proxy and the hybrid local rate-readout branch. Motivational context: the LIF and three-factor STDP equations above are included as normative biological targets and to document what a fuller circuit-level implementation would aim to reproduce.}

\section{Methods}
\subsection{Architectures}
\textbf{STDP-inspired competitive proxy (Sec.~6.4).} We implement a layered competitive pathway (Input $\rightarrow$ E1/I1 $\rightarrow$ E2/I2) with lateral inhibition and homeostatic threshold adaptation as motivation; in this paper these dynamics are abstracted by the competitive proxy described in Sec.~6.4. E1 learns unsupervised features; E2 learns decisions via reward--modulated proxy updates.

\textbf{Rate readout.} To obtain a practical, fast baseline we keep the biological encoder but average spike counts into a rate vector $\mathbf{r}$. A local, per--class delta rule adjusts a linear readout:
\begin{align}
\mathbf{p}&=\mathrm{softmax}(W\mathbf{r}+\mathbf{b}),\\
\Delta W &= \alpha\,(\mathbf{y}-\mathbf{p})\,\mathbf{r}^\top,\qquad
\Delta\mathbf{b}=\alpha\,(\mathbf{y}-\mathbf{p}).
\end{align}
This preserves local updates (pre $\times$ post) while delivering strong accuracy for benchmarking. The update rule is local in synaptic form but uses supervised labels, so it serves as a practical benchmark rather than a biologically realistic supervision mechanism.

\begin{figure}[t]
  \centering
  \includegraphics[width=0.98\linewidth]{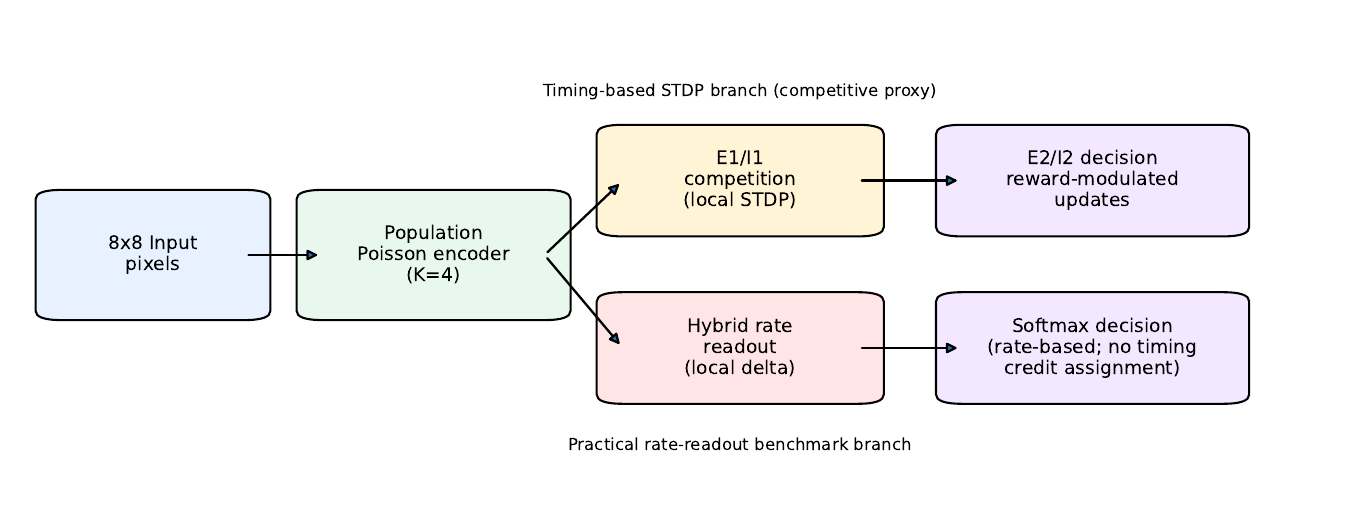}
  \caption{Schematic of the shared spiking encoder with two evaluated branches: STDP-inspired competitive proxy pathway and practical local rate-readout benchmark pathway (evaluated via proxy abstraction; no circuit simulation).}
  \label{fig:arch}
\end{figure}

\subsection{Dataset and encoding}
We use the \texttt{sklearn} digits dataset (\(8\times8\) grayscale) as a small, fast, controlled ablation platform; this is not intended as a final temporal benchmark. Intensities are normalized to $[0,1]$ and encoded with $K=4$ Gaussian--tuned Poisson neurons per pixel ($\sigma=0.25$, \(\lambda_{\max}=\)\SI{200}{Hz}). This yields a 256-dimensional encoded feature/rate vector (\(64\times4\)). Simulation step $\Delta t=\SI{1}{ms}$, stimulus window \SI{120}{ms}. Rate-window sensitivity is partially probed through $\lambda_{\max}$ ablations (100--250 Hz).
\paragraph{Deterministic sampling policy.} For each model seed, Poisson spike trains are generated per sample presentation (per epoch) using a deterministic pseudo-random generator state. This preserves stochastic sampling while keeping each seeded run exactly reproducible.

\subsection{Reward signal, stabilization, and STDP homeostasis details}
\paragraph{Reward timing and shaping.} Reward is applied once per sample at the end of the \SI{120}{ms} stimulus window. In the hybrid readout, the effective reward-gated update is implemented through the delta term
\begin{equation}
\Delta = 
\begin{cases}
\mathbf{y}-\mathbf{p}, & \text{signed reward},\\
\mathbf{y}\odot (1-\mathbf{p}), & \text{positive-only reward},
\end{cases}
\end{equation}
followed by $\Delta W=\alpha\,\Delta\,\mathbf{r}^{\top}$ and $\Delta \mathbf{b}=\alpha\,\Delta$ with $\alpha=0.003$. Signed shaping reinforces the target class and explicitly depresses competing classes through the $-\mathbf{p}$ term. Positive-only shaping reinforces only the target component, removing explicit competitor depression; this aligns with the observed lower-variance behavior in our fixed-split setting. In the STDP-style competitive proxy, signed reward is approximated by potentiating the winner neuron and depressing the runner-up prototype; positive-only removes the depressive term.
\(\odot\) denotes elementwise product.

\paragraph{Normalization and homeostasis implementation.} In the hybrid branch, we apply a post-epoch weight normalization / synaptic scaling heuristic:
\begin{equation}
W_{c,:} \leftarrow 0.98\,\frac{W_{c,:}}{\lVert W_{c,:}\rVert_2+\epsilon},
\end{equation}
with $\epsilon=10^{-8}$ and class index $c$. This operation is used as a stability heuristic and is not a full firing-rate homeostatic mechanism. In the STDP-style proxy, homeostasis refers to threshold adaptation on excitatory units:
\begin{align}
\theta_{w} &\leftarrow \theta_{w}+0.05,\\
\boldsymbol{\theta} &\leftarrow 0.995\,\boldsymbol{\theta},
\end{align}
where $w$ is the current winner neuron.

\subsection{Proxy definition and scope}
The reported STDP-inspired competitive proxy preserves local competition and bounded prototype updates but simplifies several components: (i) winner-take-all decision dynamics replace detailed recurrent E/I membrane interactions, (ii) trial-level prototype updates replace continuous conductance-based synaptic integration, and (iii) class prediction uses post-hoc neuron-to-class voting. Algorithm~\ref{alg:proxy_update} is structurally related to competitive prototype-learning and LVQ-style updates~\cite{kohonen1990self,sato1996generalized}; in this work the distinguishing elements are reward-shaped gating, threshold adaptation, and bounded post-update normalization under delayed feedback. We use this proxy to isolate the behavior of local reward and stabilization design choices under fixed controls.
Despite these simplifications, the proxy retains the central phenomenon under study: bounded local reward-modulated competition under delayed feedback.
\paragraph{Proxy update procedure.}
For each encoded sample, the proxy computes activation scores \(a = \mathbf{x}W^\top-\boldsymbol{\theta}\), selects winner and runner-up neurons, applies reward-shaped local updates to winner (potentiation) and optionally runner-up (depression for signed shaping), clips and renormalizes prototypes, updates winner threshold adaptation, and performs label lookup by neuron-to-class voting learned from training winners.
\begin{algorithm}[t]
\caption{STDP-inspired competitive proxy update (per sample)}
\label{alg:proxy_update}
\textbf{Input:} encoded sample \(\mathbf{x}\), prototypes \(W\), thresholds \(\boldsymbol{\theta}\), reward mode, \(\eta_+\), \(\eta_-\), \(w_{\min}, w_{\max}\).\\
\textbf{Step 1:} compute activation \(a \gets \mathbf{x}W^\top-\boldsymbol{\theta}\).\\
\textbf{Step 2:} choose winner \(w \gets \arg\max a\) and runner-up \(r\).\\
\textbf{Step 3:} potentiate winner prototype \(W_w \gets W_w + \eta_+(\mathbf{x}-W_w)\).\\
\textbf{Step 4:} if reward mode is signed, depress runner-up \(W_r \gets W_r-\eta_-\mathbf{x}\).\\
\textbf{Step 5:} clip and renormalize winner prototype:
\(W_w \gets \mathrm{clip}(W_w, w_{\min}, w_{\max})\),
\(W_w \gets W_w/(\lVert W_w \rVert_2+\epsilon)\).\\
\textbf{Step 6:} update thresholds \(\theta_w \gets \theta_w + \delta_\theta\), \(\boldsymbol{\theta}\gets \rho\boldsymbol{\theta}\).\\
\textbf{Step 7:} update neuron-to-class vote counts with winner \(w\) and sample label.\\
\textbf{Step 8:} infer class from winner neuron via vote lookup.
\end{algorithm}
Here \(\mathrm{clip}(\cdot)\) is an elementwise projection to \([w_{\min}, w_{\max}]\). Renormalization is per-prototype L2 normalization after clipping. When the signed-reward runner-up update is applied, the same clip and per-prototype renormalization are applied to \(W_r\); otherwise only \(W_w\) is post-processed. The runner-up depression term is used to reduce similarity to the current input and sharpen competition in this proxy setting; this is analogous to LVQ-style separation updates and is used here as a minimal local anti-correlation heuristic under delayed feedback. Clip-plus-renormalization constrains prototype magnitudes and mitigates drift/collapse, so this should be interpreted as a bounded heuristic proxy choice rather than a biophysical claim.

\subsection{Experimental protocol}
All reported baseline and ablation results use fixed seeds, explicit train/validation/test splits, and multi-seed aggregation (mean$\pm$std). We use one primary stratified split of 64\% train, 16\% validation, and 20\% test with split seed 2026. Baseline models and non-critical ablations use model seeds $\{11,23,37,41,53\}$, while the two dominant ablation axes (normalization heuristic on/off and reward shaping signed/positive-only) are extended to nine seeds $\{11,23,37,41,53,67,79,83,97\}$ to improve effect-direction confidence without rerunning the full suite. Hyperparameters are fixed a priori and reused across seeds/ablations (no per-seed retuning); the validation partition is retained to standardize future extensions and prevent test-set tuning.
In this paper, the validation partition is retained for protocol consistency and guardrails against test-set tuning; no per-factor or per-seed hyperparameters are selected via validation sweeps.

\paragraph{Robustness across dataset splits.}
To test whether the dominant hybrid effect is split-specific, we repeat key comparisons on two additional stratified splits (seeds 2027 and 2028) using the same 64/16/20 protocol. For each split we evaluate Hybrid default and Hybrid best (normalization heuristic off), holding all hyperparameters fixed and using the same model seed set $\{11,23,37,41,53\}$ without split-specific tuning. This robustness check is designed to validate effect direction, not to rerun the full benchmark suite.

\paragraph{Temporal synthetic protocol.}
To probe timing sensitivity without external data dependencies, we generate a synthetic two-class temporal-order dataset under fixed seeds and the same 64/16/20 split policy. We compare a timing-agnostic count readout against a timing-aware time-bin readout under matched local delta-rule training, each aggregated over model seeds $\{11,23,37,41,53\}$.

Raw per-seed outputs are provided in CSV form and rendered into manuscript tables through reproducible scripts.

\subsection{Key training hyperparameters}
Table~\ref{tab:key_hparams} lists the main optimization and proxy constants used in the reported runs.
\begin{table}[t]
  \centering
  \caption{Key training and proxy hyperparameters used in this study.}
  \scriptsize
  \setlength{\tabcolsep}{3pt}
  \begin{tabular}{ll}
    \toprule
    Parameter & Value \\
    \midrule
    Hybrid epochs & 18 \\
    Hybrid learning rate $\alpha$ & 0.003 \\
    Norm-on scaling / interval & 0.98 / every epoch \\
    Norm-gentle scaling / interval & 0.995 / every 5 epochs \\
    STDP-proxy epochs & 9 \\
    STDP-proxy $\eta_+$, $\eta_-$ & 0.08, 0.01 \\
    STDP-proxy bounds $(w_{\min}, w_{\max})$ & 0.0, 1.0 \\
    Threshold update $(\delta_\theta, \rho)$ & 0.05, 0.995 \\
    Timing repeats / batching & 100 / full-test vectorized \\
    \bottomrule
  \end{tabular}
  \label{tab:key_hparams}
\end{table}

\section{Results}
\subsection{Baseline results}
\IfFileExists{tables/baselines.tex}
{\begin{table}[t]
  \centering
  \caption{Baseline comparison over seeds. Accuracy is reported as mean$\pm$std, with 95\% CI half-width shown separately.}
  \scriptsize
  \setlength{\tabcolsep}{3pt}
  \resizebox{\linewidth}{!}{%
  \begin{tabular}{lccc}
    \toprule
    Method & Accuracy (mean$\pm$std, \%) & 95\% CI half-width (\%) & Macro F1 \\
    \midrule
    LogReg (pixels) & 98.06 $\pm$ 0.00 & 0.00 & 0.98 $\pm$ 0.00 \\
    MLP (pixels) & 98.22 $\pm$ 0.46 & 0.41 & 0.98 $\pm$ 0.00 \\
    LogReg (spike-encoded rates) & 96.11 $\pm$ 0.68 & 0.60 & 0.96 $\pm$ 0.01 \\
    Hybrid local readout & 86.39 $\pm$ 4.75 & 4.17 & 0.86 $\pm$ 0.06 \\
    STDP-style competitive & 87.17 $\pm$ 3.74 & 3.28 & 0.87 $\pm$ 0.04 \\
    \bottomrule
  \end{tabular}%
  }
  \label{tab:baselines}
\end{table}
}
{
\begin{table}[t]
  \centering
  \caption{Baseline comparison (see generated table output).}
  \begin{tabular}{lcc}
    \toprule
    Method & Test accuracy (\%) & Notes \\
    \midrule
    Logistic regression (pixels) & generated & classical baseline \\
    MLP (pixels) & generated & non-spiking baseline \\
    Hybrid encoder + local readout & generated & local learning rule \\
    STDP-inspired competitive proxy & generated & competitive STDP-style model \\
    \bottomrule
  \end{tabular}
  \label{tab:baselines}
\end{table}
}
Classical pixel baselines (logistic regression and MLP) exceed the current local spike-based models on this dataset; the main empirical value here is mechanistic and ablation-driven analysis rather than raw leaderboard performance. The logistic-regression row is effectively deterministic in this setup, which explains the near-zero seed variance after rounding.
The added encoded-feature logistic control is reported descriptively to separate encoder contribution from readout-rule effects under the same fixed split/seed protocol.
The LogReg (spike-encoded rates) result indicates high linear separability of the population code; the hybrid-default gap is therefore primarily associated with readout dynamics under aggressive normalization rather than encoder capacity.

\subsection{Main ablations and schedule effects}
\begin{figure}[t]
  \centering
  \includegraphics[width=0.75\linewidth]{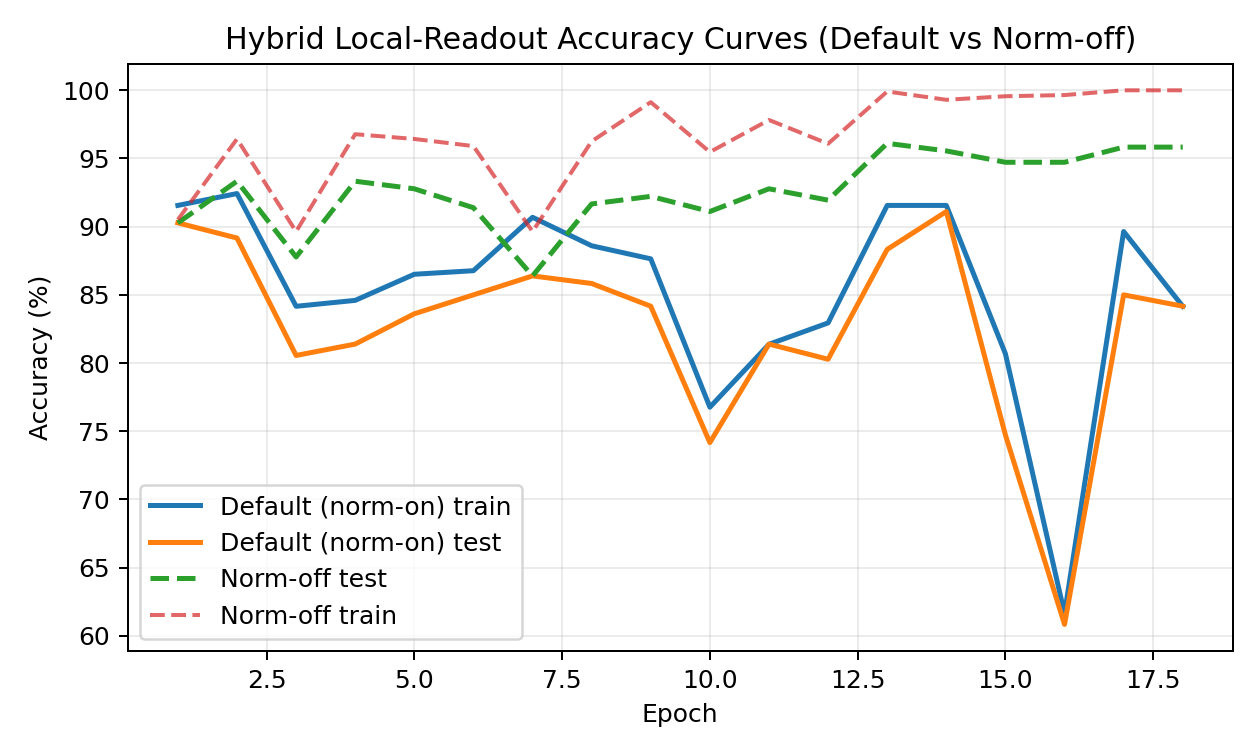}
  \caption{Accuracy trajectories for the population-coded hybrid learner, overlaying default norm-on and norm-off settings (representative model seed 23, split seed 2026).}
  \label{fig:curve}
\end{figure}

Figure~\ref{fig:curve} shows a representative training trajectory of the hybrid local-readout model. In the representative default run (model seed 23), accuracy shows a transient drop near epoch 16 before partial recovery; together with Figure~\ref{fig:norm_sched}, this is consistent with disruptive effects from aggressive per-epoch normalization relative to gentler or disabled schedules. The overlaid norm-off trajectory avoids this sharp collapse and converges more smoothly for the same split/seed.

Table~\ref{tab:ablations} reports all ablation settings, while Figure~\ref{fig:homeo_diag} visualizes the two strongest axes (normalization heuristic and reward shaping).
\IfFileExists{tables/ablations.tex}
{\begin{table}[t]
  \centering
  \caption{Ablation results over seeds. Accuracy is reported as mean$\pm$std, with 95\% CI half-width shown separately.}
  \scriptsize
  \setlength{\tabcolsep}{3pt}
  \resizebox{\linewidth}{!}{%
  \begin{tabular}{llcc}
    \toprule
    Factor & Setting & Accuracy (mean$\pm$std, \%) & 95\% CI half-width (\%) \\
    \midrule
    K & 1 & 81.50 $\pm$ 5.23 & 4.58 \\
    K & 2 & 85.78 $\pm$ 4.64 & 4.07 \\
    K & 4 & 86.39 $\pm$ 4.75 & 4.17 \\
    K & 6 & 84.28 $\pm$ 6.01 & 5.27 \\
    $\sigma$ & 0.15 & 87.44 $\pm$ 2.76 & 2.42 \\
    $\sigma$ & 0.25 & 86.39 $\pm$ 4.75 & 4.17 \\
    $\sigma$ & 0.35 & 79.33 $\pm$ 9.55 & 8.37 \\
    $\lambda_{\max}$ (Hz) & 100 & 86.67 $\pm$ 3.14 & 2.75 \\
    $\lambda_{\max}$ (Hz) & 150 & 85.22 $\pm$ 7.06 & 6.19 \\
    $\lambda_{\max}$ (Hz) & 200 & 86.39 $\pm$ 4.75 & 4.17 \\
    $\lambda_{\max}$ (Hz) & 250 & 87.50 $\pm$ 2.44 & 2.14 \\
    norm heuristic & on (n=9) & 84.44 $\pm$ 5.77 & 3.77 \\
    norm heuristic & gentle (n=9) & 88.73 $\pm$ 5.51 & 3.60 \\
    norm heuristic & off (n=9) & 95.52 $\pm$ 1.11 & 0.72 \\
    reward & signed (n=9) & 84.44 $\pm$ 5.77 & 3.77 \\
    reward & pos-only (n=9) & 91.70 $\pm$ 1.09 & 0.71 \\
    \bottomrule
  \end{tabular}%
  }
  \vspace{1mm}\\\footnotesize{Rows for normalization heuristic and reward shaping use $n=9$ seeds; all other ablation rows use $n=5$ seeds. The reward=signed row is the default configuration (normalization heuristic on), so those values match.}
  \label{tab:ablations}
\end{table}
}
{
\begin{table}[t]
  \centering
  \caption{Ablation settings used in evaluation.}
  \begin{tabular}{ll}
    \toprule
    Factor & Planned settings \\
    \midrule
    Population size $K$ & 1, 2, 4, 6 \\
    Encoder width $\sigma$ & 0.15, 0.25, 0.35 \\
    Peak rate $\lambda_{\max}$ & 100, 150, 200, 250 Hz \\
    Norm heuristic & on/gentle/off \\
    Reward shaping & positive-only vs signed reward \\
    \bottomrule
  \end{tabular}
  \label{tab:ablations}
\end{table}
}
Table~\ref{tab:baselines} reports expected pixel-baseline performance (98.06$\pm$0.00\% and 98.22$\pm$0.46\%) versus lower local spike-based baselines (86.39$\pm$4.75\% and 87.17$\pm$3.74\%); the 98.06$\pm$0.00 value reflects rounding to two decimals. Table~\ref{tab:ablations} shows the dominant levers: disabling the post-epoch normalization heuristic reaches 95.52$\pm$1.11\%, while the gentle schedule reaches 88.73$\pm$5.51\%, and the positive-only setting in the norm-on paired context reaches 91.70$\pm$1.09\%.

\subsection{Norm-reward interaction}
\begin{figure}[t]
  \centering
  \includegraphics[width=0.95\linewidth]{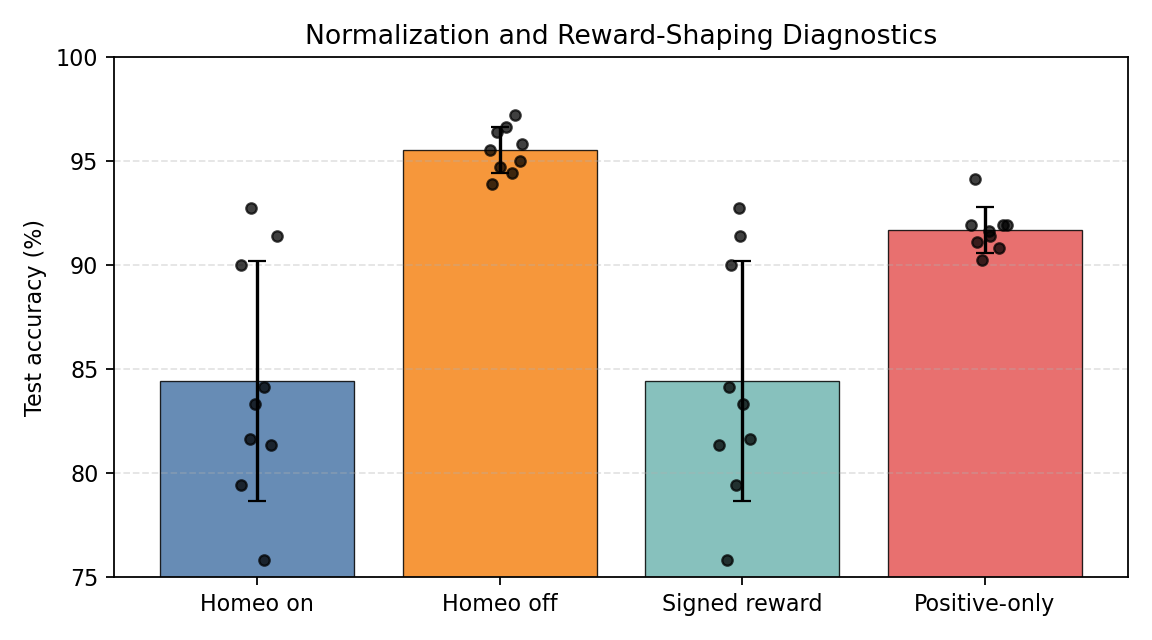}
  \caption{Seed-level diagnostic summary for normalization-heuristic and reward-shaping ablations (bars: mean$\pm$std; points: per-seed values for seeds \{11,23,37,41,53,67,79,83,97\}).}
  \label{fig:homeo_diag}
\end{figure}

\IfFileExists{tables/norm_reward_2x2.tex}{\begin{table}[t]
  \centering
  \caption{Explicit 2$\times$2 interaction summary for normalization and reward shaping on \texttt{sklearn} digits (descriptive, no factorial significance claim).}
  \scriptsize
  \setlength{\tabcolsep}{3pt}
  \resizebox{\linewidth}{!}{%
  \begin{tabular}{lccc}
    \toprule
    Condition & Accuracy (mean$\pm$std, \%) & 95\% CI half-width (\%) & Notes \\
    \midrule
    norm on + reward signed & 84.44 $\pm$ 5.77 & 3.77 & default \\
    norm on + reward pos-only & 91.70 $\pm$ 1.09 & 0.71 & targeted run \\
    norm off + reward signed & 95.52 $\pm$ 1.11 & 0.72 & ablation row \\
    norm off + reward pos-only & 91.98 $\pm$ 1.49 & 0.97 & targeted run \\
    \midrule
    $\Delta$ pos-only$-$signed (norm on) & 7.25 & -- & descriptive \\
    $\Delta$ pos-only$-$signed (norm off) & -3.55 & -- & descriptive \\
    \bottomrule
  \end{tabular}%
  }
  \vspace{1mm}\\\footnotesize{$n=9$ seeds in each cell, using the dominant-axis seed set.}
  \label{tab:norm_reward_2x2}
\end{table}
}{}
Table~\ref{tab:norm_reward_2x2} gives an explicit descriptive 2$\times$2 view of normalization and reward-shaping settings. This table is reported as interaction context only and is not used as a formal factorial significance claim. Figure~\ref{fig:homeo_diag} summarizes seed-level accuracy distributions for the two strongest ablation axes.

\subsection{Mechanism diagnostics}
\begin{figure}[t]
  \centering
  \includegraphics[width=0.95\linewidth]{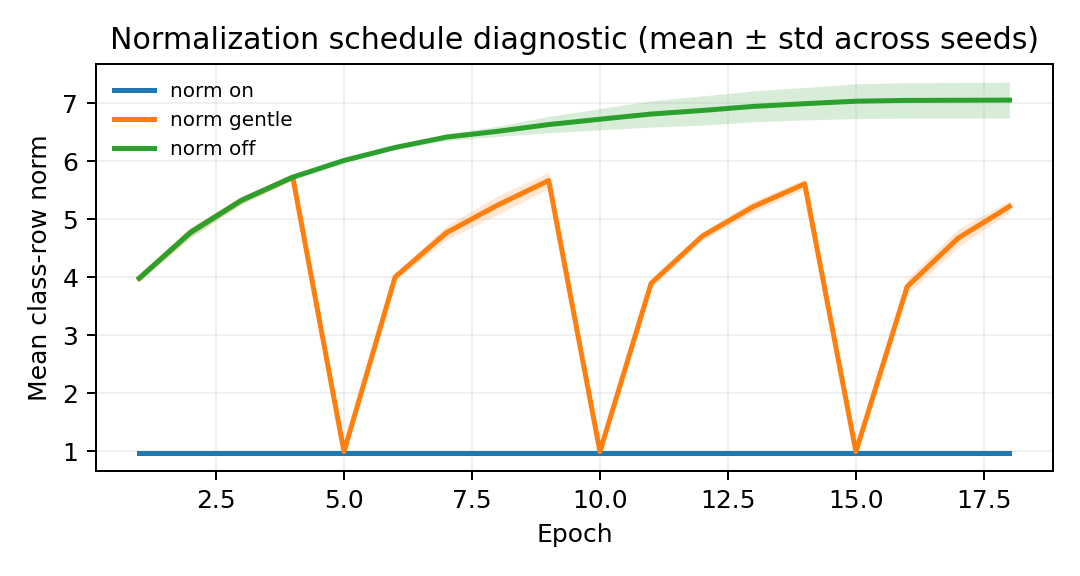}
  \caption{Normalization-schedule diagnostic: mean class-row norm trajectory across training epochs (mean$\pm$std over seeds).}
  \label{fig:norm_sched}
\end{figure}

Figure~\ref{fig:norm_sched} reports mean class-row norm trajectories for normalization on, gentle, and off schedules; the y-axis is the mean L2 norm of class-row weights \(W_{c,:}\) over epochs. The gentle schedule sits between aggressive per-epoch normalization and fully disabled normalization, supporting the interpretation that schedule aggressiveness (not merely presence/absence of stabilization) drives the dominant accuracy shift. The gentle curve shows a sawtooth pattern because normalization is applied every five epochs, allowing norm growth between scheduled rescaling steps.

\subsection{Paired significance checks}
For key seed-matched comparisons, we report exploratory two-sided exact sign tests, paired effect-size confidence intervals, and paired effect sizes (\(d_z\), Cliff's delta). Seeds are treated as the unit of replication. Hybrid versus STDP-style baseline difference remains small (mean $-0.78$ percentage points, sign test $n=5$, $p=1.0000$, paired CI$_{95}$ half-width $\pm4.31$ pp, \(d_z=-0.16\), Cliff's delta $=-0.20$). Normalization-heuristic off versus on in the hybrid branch is associated with the largest observed improvement (mean $+11.08$ pp, sign test $n=9$, $p=0.0039$, paired CI$_{95}$ half-width $\pm3.28$ pp, \(d_z=2.21\), Cliff's delta $=1.00$). Reward shaping interacts with stabilization: under normalization-on, positive-only improves mean accuracy relative to signed reward (mean $+7.25$ pp over all nine paired seeds, with $n=8$ non-tied pairs; paired CI$_{95}$ half-width $\pm3.83$ pp over nine pairs; \(d_z=1.24\), Cliff's delta $=0.67$), while Table~\ref{tab:norm_reward_2x2} shows this contrast reverses under normalization-off. Given the small sample regime, p-values are exploratory; confidence intervals, effect sizes, and effect direction are treated as the primary evidence.
We do not apply multiple-comparisons correction, as these tests are reported descriptively to summarize seed-level stability rather than to make confirmatory significance claims.

\subsection{Robustness across dataset splits}
\IfFileExists{tables/split_robustness.tex}
{\begin{table}[t]
  \centering
  \caption{Robustness across dataset splits (no retuning). Per-split values aggregate over five model seeds; $\Delta$ denotes best minus default. Reward shaping is signed in both compared conditions; only the normalization heuristic is toggled.}
  \scriptsize
  \setlength{\tabcolsep}{3pt}
  \resizebox{\linewidth}{!}{%
  \begin{tabular}{lccc}
    \toprule
    Split seed & Hybrid default Acc (mean$\pm$std, \%) & Hybrid best (norm off) Acc (mean$\pm$std, \%) & $\Delta$ (pp) \\
    \midrule
    2026 & 86.39 $\pm$ 4.75 & 95.50 $\pm$ 0.84 & +9.11 \\
    2027 & 85.39 $\pm$ 4.08 & 93.06 $\pm$ 0.81 & +7.67 \\
    2028 & 81.50 $\pm$ 7.58 & 92.67 $\pm$ 1.83 & +11.17 \\
    \midrule
    Across splits & 84.43 $\pm$ 2.58 & 93.74 $\pm$ 1.54 & +9.31 $\pm$ 1.76 (\(\Delta>0\) in 3/3) \\
    \bottomrule
  \end{tabular}%
  }
  \vspace{1mm}\\\footnotesize{Per-split means are computed over $n=5$ model seeds; across-split summaries are computed over $n=3$ split-level means. Dominant-axis ablations in Table~\ref{tab:ablations} use $n=9$ seeds.}
  \label{tab:split_robustness}
\end{table}
}
{}
Table~\ref{tab:split_robustness} shows that the normalization-off improvement is preserved across split seeds under fixed hyperparameters and no retuning, with \(\Delta>0\) in all three splits, reducing the risk that the dominant ablation effect is an artifact of a single partition.
Table~\ref{tab:split_robustness} uses the 5-seed split-robustness set, whereas Table~\ref{tab:ablations} reports dominant-axis ablations over 9 seeds; small numeric differences across these summaries are therefore expected.

\subsection{Temporal synthetic benchmark}
\IfFileExists{tables/temporal_synthetic.tex}
{\begin{table}[t]
  \centering
  \caption{Temporal synthetic benchmark (fixed seeds). Accuracy is reported as mean$\pm$std, with 95\% CI half-width shown separately.}
  \scriptsize
  \setlength{\tabcolsep}{3pt}
  \resizebox{\linewidth}{!}{%
  \begin{tabular}{lccc}
    \toprule
    Condition & Accuracy (mean$\pm$std, \%) & 95\% CI half-width (\%) & Macro F1 \\
    \midrule
    Count readout local (timing-agnostic) & 50.25 $\pm$ 0.96 & 0.84 & 0.37 $\pm$ 0.06 \\
    Time-bin readout local (timing-aware) & 84.62 $\pm$ 1.22 & 1.07 & 0.85 $\pm$ 0.01 \\
    \bottomrule
  \end{tabular}%
  }
  \vspace{1mm}\\\footnotesize{$n=5$ seeds per condition; synthetic temporal-order classification with fixed split seed 2026.}
  \label{tab:temporal_synth}
\end{table}
}
{}
The synthetic temporal-order task separates timing-agnostic from timing-aware local readouts under matched optimization, providing a network-free stress test for the timing-credit limitation discussed for static count readouts.
On this task, count-only local readout remains near chance (50.25$\pm$0.96\%), while the timing-aware time-bin local readout reaches 84.62$\pm$1.22\%.

\IfFileExists{tables/openml_benchmark.tex}{
\subsection{External OpenML benchmark}
\begin{table}[t]
  \centering
  \caption{OpenML external benchmark on MNIST (fixed split/seed protocol). Accuracy is reported as mean$\pm$std, with 95\% CI half-width shown separately.}
  \scriptsize
  \setlength{\tabcolsep}{3pt}
  \resizebox{\linewidth}{!}{%
  \begin{tabular}{lccc}
    \toprule
    Condition & Accuracy (mean$\pm$std, \%) & 95\% CI half-width (\%) & Macro F1 \\
    \midrule
    Hybrid default (norm on, signed) & 79.28 $\pm$ 7.83 & 6.87 & 0.79 $\pm$ 0.08 \\
    Hybrid best (norm off, signed) & 85.03 $\pm$ 1.82 & 1.59 & 0.85 $\pm$ 0.02 \\
    Hybrid norm on (pos-only) & 84.60 $\pm$ 0.22 & 0.20 & 0.84 $\pm$ 0.00 \\
    Hybrid norm off (pos-only) & 84.60 $\pm$ 2.15 & 1.88 & 0.85 $\pm$ 0.02 \\
    STDP proxy (norm on, signed) & 57.34 $\pm$ 4.37 & 3.83 & 0.55 $\pm$ 0.07 \\
    LogReg (spike-encoded rates) & 88.32 $\pm$ 0.14 & 0.12 & 0.88 $\pm$ 0.00 \\
    \midrule
    $\Delta$ (norm off $-$ norm on, signed) & 5.75 & -- & -- \\
    $\Delta$ (pos-only $-$ signed, norm on) & 5.32 & -- & -- \\
    $\Delta$ (pos-only $-$ signed, norm off) & -0.44 & -- & -- \\
    \bottomrule
  \end{tabular}%
  }
  \vspace{1mm}\\\footnotesize{$n=5$ seeds per condition; split seed 2026; no per-split tuning. The STDP proxy row is a controlled abstraction and is not presented as a competitive MNIST SNN baseline.}
  \label{tab:openml_bench}
\end{table}

The OpenML benchmark extends evaluation beyond \texttt{sklearn} digits and checks whether dominant effect direction is preserved on a larger external dataset under the same fixed-seed, no-retuning protocol. We now report the hybrid 2$\times$2 cells (norm on/off $\times$ signed/pos-only), the STDP proxy branch, and an encoded-feature logistic control under the same split/seed contract.
The dominant normalization effect direction transfers to MNIST with smaller magnitude (norm-off minus norm-on signed), indicating direction-level generalization but dataset-dependent effect size. The encoded-feature logistic row provides encoder-separability context on MNIST, while the hybrid default gap remains associated with readout dynamics under aggressive normalization.
The MNIST 2$\times$2 interaction is reported descriptively in Table~\ref{tab:openml_norm_reward_2x2}; as with digits, reward-shaping interpretation is read jointly with stabilization regime rather than as a standalone main effect.
On MNIST, the interaction takes a different form: positive-only outperforms signed under norm-on (+5.32 pp), while the two are nearly equivalent under norm-off (-0.44 pp for pos-only minus signed), reinforcing that reward-shaping behavior is jointly regime- and dataset-dependent.
This cross-dataset pattern supports the core thesis: the interaction direction is robust under fixed local-update constraints, while the effect magnitude remains dataset-dependent.
The low standard deviation of the norm-on positive-only row is an observed outcome under the same deterministic seed protocol used across conditions and is shown directly in the seed-level appendix values (Table~\ref{tab:openml_seed_values}).
The large spread of the MNIST norm-on signed cell (79.28$\pm$7.83\%) is driven by a single low run (seed 53: 66.32\%), consistent with occasional instability under aggressive normalization at this scale.
We additionally verified seed diversity by confirming non-identical seed-specific spike-count aggregates and encoded trajectories across runs, so the low spread is interpreted as stable convergence rather than duplicated sampling.
For context, this MNIST range remains below reported unsupervised STDP benchmarks such as Diehl and Cook (2015, \(\sim95\%\) on MNIST). Their setup uses full recurrent E/I dynamics with unsupervised STDP and different preprocessing/encoder assumptions; our proxy intentionally abstracts those circuit dynamics to isolate interaction effects under a fixed encoder, so the gap is interpreted as a quantifiable cost of this controlled simplification and a target for fuller recurrent extensions.
}{}

\IfFileExists{tables/openml_norm_reward_2x2.tex}{
\begin{table}[t]
  \centering
  \caption{OpenML MNIST descriptive 2$\times$2 interaction context for hybrid reward shaping and normalization regime (accuracy, \%).}
  \scriptsize
  \setlength{\tabcolsep}{4pt}
  \begin{tabular}{lc}
    \toprule
    Cell & Accuracy (mean$\pm$std) \\
    \midrule
    Norm on + signed & 79.28 $\pm$ 7.83 \\
    Norm on + pos-only & 84.60 $\pm$ 0.22 \\
    Norm off + signed & 85.03 $\pm$ 1.82 \\
    Norm off + pos-only & 84.60 $\pm$ 2.15 \\
    \midrule
    $\Delta$ (pos-only $-$ signed | norm on) & 5.32 $\pm$ 7.73 \\
    $\Delta$ (pos-only $-$ signed | norm off) & -0.44 $\pm$ 2.65 \\
    \bottomrule
  \end{tabular}
  \vspace{1mm}\\\footnotesize{Descriptive interaction context only ($n=5$ seeds/cell); no confirmatory factorial significance claim.}
  \label{tab:openml_norm_reward_2x2}
\end{table}

}{}

\IfFileExists{tables/openml_seed_values.tex}{
\begin{table}[t]
  \centering
  \caption{OpenML MNIST per-seed accuracy values (\%) for the six reported conditions (split seed 2026).}
  \scriptsize
  \setlength{\tabcolsep}{3pt}
  \resizebox{\linewidth}{!}{%
  \begin{tabular}{lccccc}
    \toprule
    Condition & seed 11 & seed 23 & seed 37 & seed 41 & seed 53 \\
    \midrule
    Hybrid default (norm on, signed) & 79.435714 & 84.907143 & 79.692857 & 86.042857 & 66.321429 \\
    Hybrid best (norm off, signed) & 85.035714 & 87.607143 & 83.442857 & 85.885714 & 83.200000 \\
    Hybrid norm on (pos-only) & 84.392857 & 84.871429 & 84.807143 & 84.514286 & 84.421429 \\
    Hybrid norm off (pos-only) & 80.828571 & 85.721429 & 85.664286 & 85.914286 & 84.864286 \\
    STDP proxy (norm on, signed) & 61.478571 & 58.807143 & 60.971429 & 51.864286 & 53.592857 \\
    LogReg (spike-encoded rates) & 88.164286 & 88.442857 & 88.421429 & 88.392857 & 88.164286 \\
    \bottomrule
  \end{tabular}%
  }
  \vspace{1mm}\\\footnotesize{Values are taken directly from \texttt{results/openml\_benchmark\_raw.csv}; no additional runs.}
  \label{tab:openml_seed_values}
\end{table}

}{}

\subsection{Diagnostic visualizations}
\begin{figure}[t]
  \centering
  \includegraphics[width=0.75\linewidth]{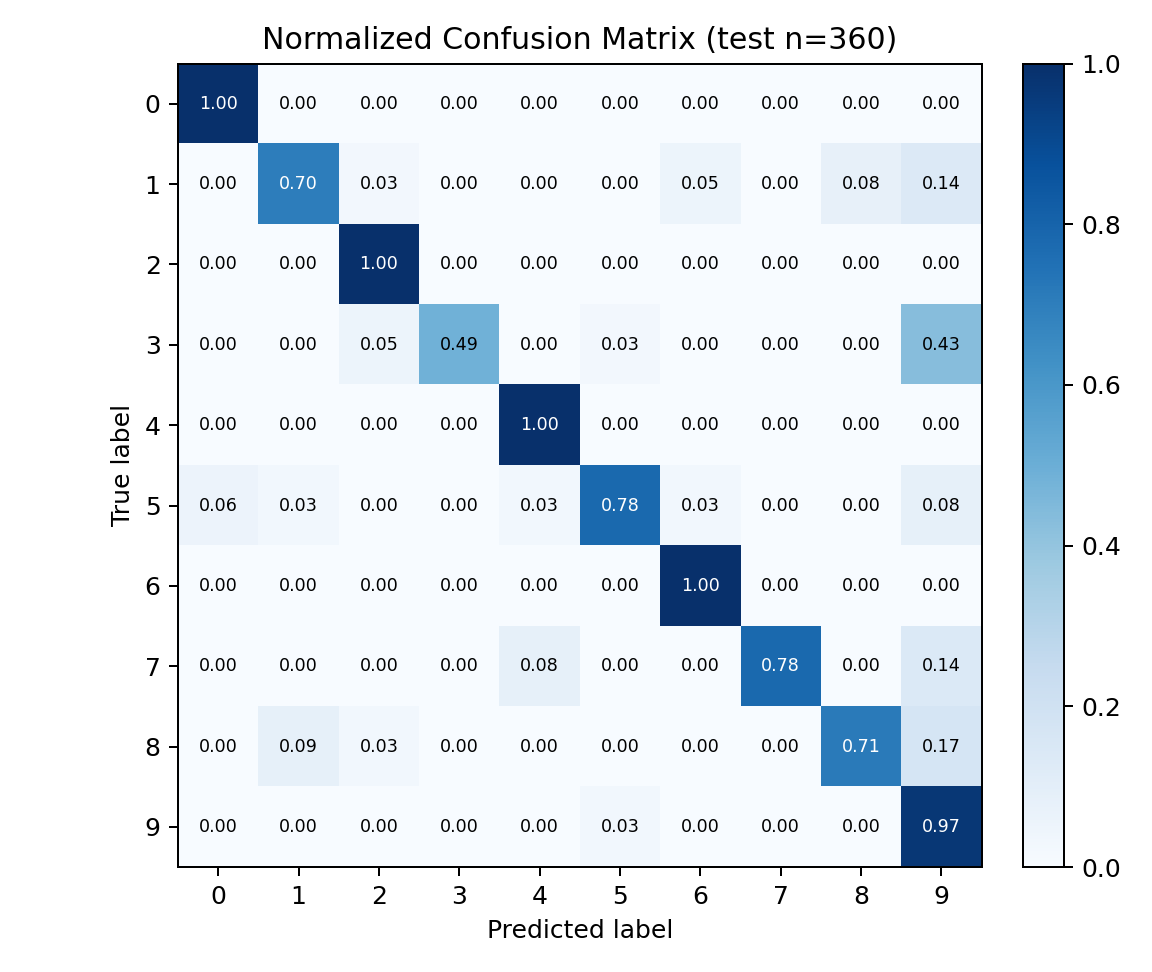}
  \caption{Normalized confusion matrix on the held--out test set (representative model seed 23, split seed 2026).}
  \label{fig:cm}
\end{figure}

\begin{figure}[t]
  \centering
  \includegraphics[width=0.95\linewidth]{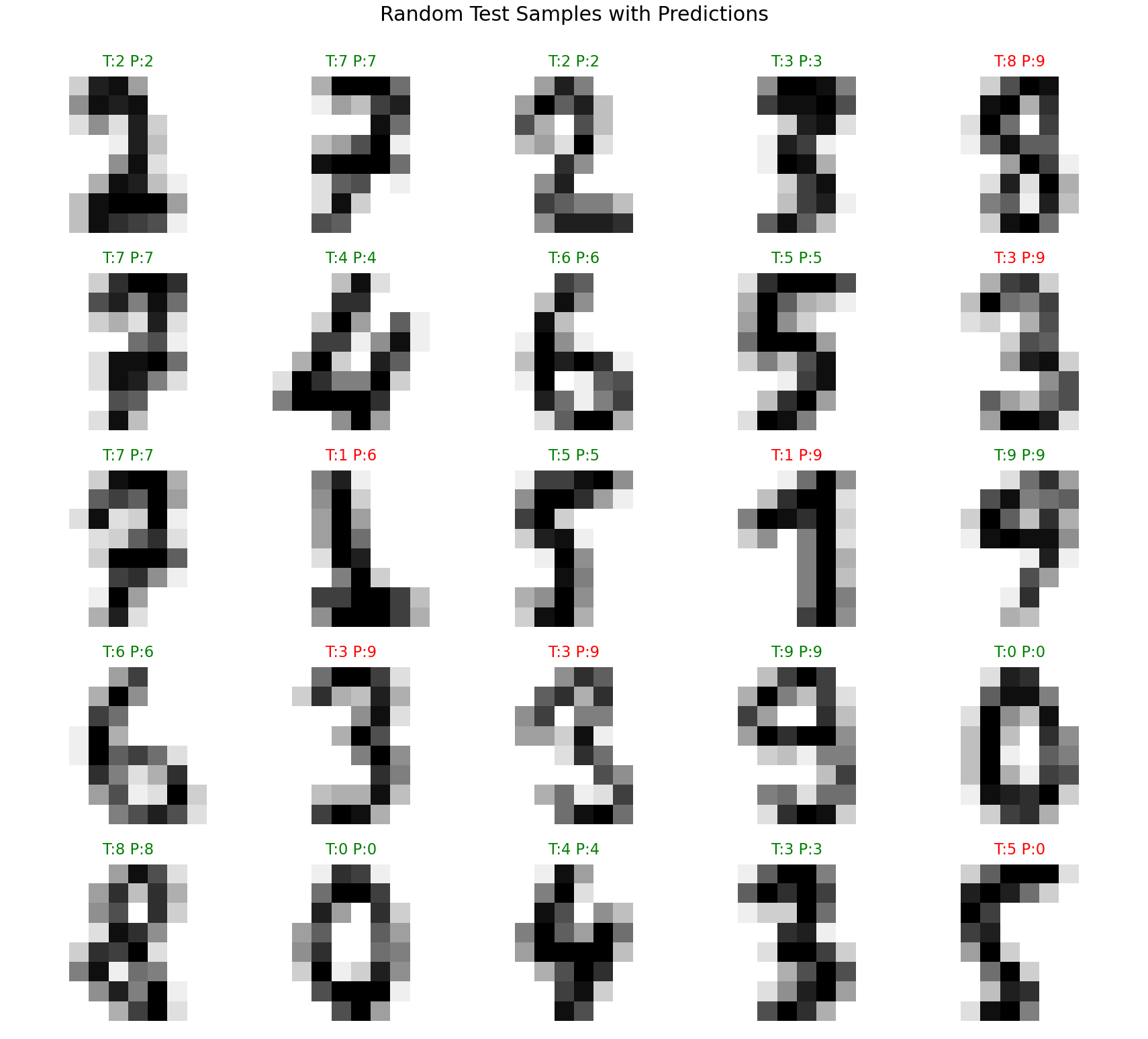}
  \caption{Random test digits with predicted labels (representative model seed 23, split seed 2026).}
  \label{fig:samples}
\end{figure}

Figure~\ref{fig:cm} shows the normalized confusion matrix for the hybrid local-readout model, and Figure~\ref{fig:samples} shows qualitative sample predictions on held-out test images.

\subsection{Additional SNN metrics}
\IfFileExists{tables/extra_metrics.tex}
{\begin{table}[t]
  \centering
  \caption{Additional SNN diagnostics (hybrid default; STDP rows explicitly marked). Inference timing is reported as median over 100 repeats on full-test vectorized batches ($N=360$): forward-only excludes encoding and I/O, while end-to-end includes encoding and forward pass; no learning/update steps are included. Parameter counts include model weights and threshold terms; STDP-style counts also include excitatory prototype vectors ($96\times256$).}
  \scriptsize
  \setlength{\tabcolsep}{3pt}
  \resizebox{\linewidth}{!}{%
  \begin{tabular}{lc}
    \toprule
    Metric & Value \\
    \midrule
    Hybrid default: worst class F1 (digit 8) & 0.63 \\
    Hybrid default: 2nd worst class F1 (digit 9) & 0.74 \\
    Hybrid default: input spikes/sample (mean $\pm$ std across seeds) & 2420.7 $\pm$ 2.4 \\
    STDP branch: weight saturation (\% at lower bound) & 2.55 $\pm$ 0.15 \\
    STDP branch: weight saturation (\% at upper bound) & 0.00 $\pm$ 0.00 \\
    STDP branch: winner margin (activation units, normalized dot-product) & 0.0229 $\pm$ 0.0014 \\
    Hybrid parameter count & 2570 \\
    STDP-style parameter count & 24672 \\
    Hybrid timing: forward-only median ($\mu$s/sample, 100 repeats) & 0.160 \\
    Hybrid timing: end-to-end median ($\mu$s/sample, 100 repeats) & 8.773 \\
    STDP timing: forward-only median ($\mu$s/sample, 100 repeats) & 0.653 \\
    STDP timing: end-to-end median ($\mu$s/sample, 100 repeats) & 9.410 \\
    \bottomrule
  \end{tabular}%
  }
  \label{tab:extra_metrics}
\end{table}
}
{
\begin{table}[t]
  \centering
  \caption{Additional SNN diagnostics.}
  \begin{tabular}{lc}
    \toprule
    Metric & Value \\
    \midrule
    Worst class F1 & generated \\
    Input spikes/sample & generated \\
    Parameter count & generated \\
    \bottomrule
  \end{tabular}
  \label{tab:extra_metrics}
\end{table}
}
The STDP saturation and winner-margin diagnostics indicate non-trivial competitive sparse-coding behavior rather than a random-readout artifact in the proxy branch.
Winner margin is defined as \(a_{\text{winner}} - a_{\text{runner-up}}\), averaged over test samples, where \(a\) denotes the STDP branch activation score before label lookup.
Here \(a\) is the normalized prototype-similarity score (\(\mathbf{x}\mathbf{w}^{\top}-\theta\)) used by the competitive proxy before label lookup, with both encoded features \(\mathbf{x}\) and prototype weights \(\mathbf{w}\) L2-normalized and \(\theta\) denoting the neuron threshold term.
In the representative confusion matrix (Figure~\ref{fig:cm}; split seed 2026, model seed 23), the largest error concentration is digit 3 being predicted as 9. Table~\ref{tab:extra_metrics} reports seed-aggregated per-class F1 summaries, so the worst-class ranking can differ from any single-seed confusion snapshot.
The low spike-count variance is expected because each sample aggregates many Poisson events (pixels \(\times\) population channels \(\times\) time bins), which reduces relative variance.
Inference timing uses \verb|time.perf_counter| on Apple Silicon macOS arm64 (CPU model: Apple M4 Pro; Python 3.13.5) with a full-test-set vectorized batch, reporting median per-sample latency over 100 repeats. We report both forward-only timing (excludes encoding and data loading) and end-to-end timing (includes encoding and forward pass, excludes data loading). Inference-only; no learning/update steps are included.
Reported per-sample timings are amortized values from vectorized batch timing (\(t_{\text{batch}}/N\)), not streaming per-sample latency.

\section{Design Recommendations}
Based on the controlled ablations and external checks, we recommend: (i) treat normalization schedule aggressiveness as a first-order design variable and evaluate on/gentle/off schedules explicitly; (ii) report reward-shaping conclusions jointly with stabilization regime using an explicit 2$\times$2 summary; (iii) include encoded-feature logistic controls to separate encoder capacity from readout-rule dynamics; and (iv) use timing-aware readouts when task structure is temporally coded, as count-only readouts can fail near chance despite matched local updates.

\section{Limitations}
Local spike-based models in this study remain below classical pixel baselines on \texttt{sklearn} digits. The implemented STDP result is a competitive proxy model rather than a full biophysical E/I simulator. The full LIF/three-factor equations are included as theoretical motivation, while the evaluated implementation is Algorithm~\ref{alg:proxy_update}; this separation should be interpreted explicitly. Our encoder and local updates are motivated by biologically inspired learning hypotheses (Poisson population spikes, local activity terms, bounded updates, and competition), but the practical readout uses supervised per-trial labels and discards precise spike timing. This timing-credit limitation is corroborated by the synthetic temporal benchmark (Table~\ref{tab:temporal_synth}), where count-only readout fails on temporally structured inputs. We validate normalization and reward-shaping interaction behavior on MNIST under fixed seeds, but broader robustness over additional real-world datasets and temporal tasks remains future work. This submission prioritizes controlled interaction evidence under fixed seeds over full benchmark coverage, and the released harness is intended to support systematic extension to additional datasets. Achieving parity likely requires richer recurrent credit assignment and tighter hardware-constrained evaluation beyond controlled benchmarks.

\section{Reproducibility}
All quantitative claims in this paper are backed by fixed-seed CSV outputs, generated tables, generated figures, and scripted build/validation steps. The submission bundle includes manuscript sources, generation scripts, QA/signoff scripts, and checksum verification. The manuscript source auto-compiles with IEEEtran when available and uses a compatible two-column fallback otherwise.
Code and data artifacts are also available at \url{https://github.com/spoutop/spiking-local-learning-benchmarks}.

\section{Conclusion}
Across fixed-seed benchmarks, normalization-schedule aggressiveness is the dominant factor controlling local readout performance, and it modulates the direction of reward-shaping effects. The synthetic timing-coded task shows count-based readouts near chance while timing-aware readouts succeed, supporting the timing-versus-rate limitation interpretation under matched local updates.
Across the reported ablations and external checks, reward-shaping effects are regime- and dataset-dependent and should be interpreted jointly with stabilization settings. Overall, these results support a controlled benchmark interpretation: stabilization schedule is a first-order design variable, and timing-coded tasks require timing-aware readouts.

\section*{Acknowledgment}
We thank the open--source community of neuromorphic researchers for foundational ideas.

\end{document}